\documentclass[conference]{IEEEtran}
\IEEEoverridecommandlockouts
\usepackage{cite}
\usepackage{amsmath,amssymb,amsfonts}
\usepackage{algorithmic}
\usepackage{graphicx}
\usepackage{textcomp}
\usepackage{xcolor}
\usepackage{booktabs} 
\usepackage{siunitx}  
\def\BibTeX{{\rm B\kern-.05em{\sc i\kern-.025em b}\kern-.08em
    T\kern-.1667em\lower.7ex\hbox{E}\kern-.125emX}}

\begin{document}

\title{CT Scans As Video: Efficient Intracranial Hemorrhage Detection Using Multi-Object Tracking}

\author{
Amirreza Parvahan\textsuperscript{a}, 
Mohammad Hoseyni\textsuperscript{b},
Javad Khoramdel\textsuperscript{c},
and Amirhossein Nikoofard\textsuperscript{b*}\thanks{*Corresponding author}\\\\
\textsuperscript{a}\textit{Faculty of Computer Engineering, K. N. Toosi University of Technology, Tehran, Iran} \\
\textsuperscript{b}\textit{Faculty of Electrical Engineering, K. N. Toosi University of Technology, Tehran, Iran}\\
\textsuperscript{c}\textit{Faculty of Mechanical Engineering, Tarbiat Modares University, Tehran, Iran} \\
\textit{\{amirrezaparv, mohammadhosini60, j.khorramdel96\}@gmail.com, a.nikoofard@kntu.ac.ir}
}

\maketitle

\begin{figure*}[!t]
\centering
\includegraphics[width=\textwidth, 
                 trim={0 200pt 0 150pt}, 
                 clip]{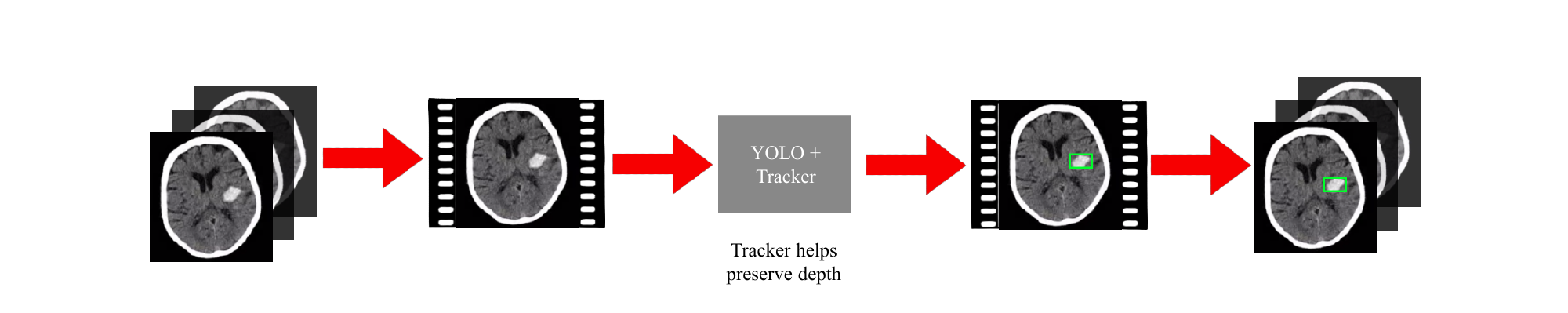} 
\caption{Overview of the proposed video-based detection pipeline.}
\label{fig:pipeline} 
\end{figure*}

\begin{abstract}
Automated analysis of volumetric medical imaging on edge devices is severely constrained by the high memory and computational demands of 3D Convolutional Neural Networks (CNNs). This paper develops a lightweight computer vision framework that reconciles the efficiency of 2D detection with the necessity of 3D context by reformulating volumetric Computer Tomography (CT) data as sequential video streams. This video-viewpoint paradigm is applied to the time-sensitive task of Intracranial Hemorrhage (ICH) detection using the Hemorica dataset. To ensure operational efficiency, we benchmarked multiple generations of the YOLO architecture (v8, v10, v11 and v12) in their Nano configurations, selecting the version with the highest mAP@50 to serve as the slice-level backbone. A ByteTrack algorithm is then introduced to enforce anatomical consistency across the $z$-axis. To address the initialization lag inherent in video trackers, a hybrid inference strategy and a spatiotemporal consistency filter are proposed to distinguish true pathology from transient prediction noise. Experimental results on independent test data demonstrate that the proposed framework serves as a rigorous temporal validator, increasing detection Precision from 0.703 to 0.779 compared to the baseline 2D detector, while maintaining high sensitivity. By approximating 3D contextual reasoning at a fraction of the computational cost, this method provides a scalable solution for real-time patient prioritization in resource-constrained environments, such as mobile stroke units and IoT-enabled remote clinics.
\end{abstract}


\section{Introduction}
Modern service-oriented environments, particularly in critical healthcare, face a systemic bottleneck that compromises operational efficiency and user outcomes: service latency. The combination of high data influx with CT utilization in emergency departments increasing exponentially over the last few decades~\cite{Larson2011} and a persistent shortage of specialist experts creates a dangerous queue for critical decision-making. This wait time represents a high-risk window where a subject's condition can rapidly deteriorate while awaiting expert analysis. The urgency of this optimization challenge is particularly acute in the detection of Intracranial Hemorrhage (ICH), where diagnostic latency directly correlates with irreversible neurological injury and mortality~\cite{greenberg2022guideline, Coles2007}. In these high-stakes scenarios, the primary goal of an AI system is process optimization: to function as an automated triage agent that prioritizes cases based on urgency, thereby minimizing the time-to-intervention for the most critical patients~\cite{Papa2012}.

In current clinical practice, experts (radiologists) do not analyze data as static, isolated snapshots. When reviewing a Computed Tomography (CT) scan, they scroll through sequential slices, mentally reconstructing the 3D anatomy to distinguish true anomalies from noise. This mental process of tracking a lesion across the $z$-axis allows them to validate continuity and shape. However, automating this workflow on standard hardware presents a significant engineering challenge. Cranial fractures and hemorrhages, particularly when coursing in the axial plane, remain some of the most commonly missed major abnormalities on head CT scans due to human fatigue and interpretation speed~\cite{Wysoki1998, Erly2002}.

The selection of object detection as the primary task as opposed to segmentation or simple classification is driven by distinct operational and reliability constraints. While semantic segmentation provides granular detail, it demands labor-intensive pixel-wise annotation and incurs a computational overhead often prohibitive for real-time edge deployment. Conversely, while binary classification has been successfully applied to head CTs~\cite{Gao2017, Grewal2017}, it inherently lacks spatial interpretability; the model makes a prediction without explicitly defining the anatomical region of interest. To identify the features driving a classification decision, one must rely on post-hoc Explainable AI (XAI) techniques~\cite{rafati2025benchmarking}. However, unlike classification where the attention mechanism is not explicitly defined in the training objective, object detection enforces regional-based learning. By supervising the network with bounding boxes, the model is directly constrained to focus on relevant anatomical features, ensuring diagnostic reliability is built into the learning process rather than interpreted afterwards.

Existing automated solutions represent two extremes of the computational spectrum. 
The first, slice-based 2D detection, often utilizing real-time architectures like 
YOLO \cite{Redmon_2016_CVPR}, treats every data frame as an independent event. While these models are lightweight enough for real-time edge deployment, they suffer from temporal amnesia lacking awareness of the context in preceding or succeeding frames which leads to false positives. The second paradigm, 3D volumetric learning, captures full spatial context but demands massive computational resources and memory bandwidth~\cite{Litjens2017}. This reliance on heavy compute makes 3D models impractical for Edge AI scenarios such as mobile clinics or IoT-enabled scanners where hardware resources are strictly constrained and low latency is paramount.

There is a distinct need for a middle ground that combines the precision of contextual analysis with the efficiency of 2D inference. This paper explores the concept of tracking-by-detection, a technique borrowed from video analytics, applied here to optimize medical image processing. By viewing the $z$-axis of a CT scan as a temporal sequence, lightweight object tracking algorithms can be employed to maintain the identity of a lesion across slices. \textbf{To the best of our knowledge, this work represents the first attempt to explicitly formulate medical lesion detection as a video object tracking problem.} By treating the volume as a video stream, we aim to bridge the gap between 2D efficiency and 3D context without the computational penalty of volumetric networks.

The contributions of this study are summarized as follows:
\begin{enumerate}
    \item A Novel Video-Viewpoint Framework: We propose a paradigm shift in volumetric analysis by reformulating CT data as sequential video streams. This approach bridges the gap between 2D efficiency and 3D context, enabling volumetric reasoning on resource-constrained edge devices without the computational overhead of 3D CNNs.
    \item Hybrid Tracking Strategy: We identify and resolve the initialization lag inherent in standard video trackers (e.g., ByteTrack) when applied to static medical volumes. A novel Hybrid Inference strategy and a Spatiotemporal Consistency Filter are introduced, effectively fusing the high sensitivity of slice-based detectors with the temporal consistency of motion trackers.
\end{enumerate}

The remainder of this paper is structured as follows. Section II reviews prior research on intracranial hemorrhage detection, dataset evolution, and the application of object tracking in medical imaging. Section III details the proposed methodology, covering the preprocessing of the Hemorica dataset, the YOLOv11n backbone, and the specific adaptations required for the ByteTrack algorithm. Section IV presents the experimental setup, hyperparameter optimization, and a comprehensive analysis of the results. Section V discusses the clinical implications and limitations of the findings. Finally, Section VI concludes the study and outlines future research directions.

\section{Related Works}
The automation of intracranial hemorrhage detection has evolved from simple slice-level classification to complex volumetric segmentation. However, a persistent challenge remains: balancing the high computational cost of 3D context with the efficiency required for clinical deployment. This section reviews the progression of datasets, detection methodologies, and the emerging role of object tracking in medical image analysis.

\subsection{Datasets for Intracranial Hemorrhage}
The development of automated ICH detection has been heavily influenced by the availability of public datasets. Early benchmarks like PhysioNet~\cite{hssayeni2020intracranial} provided limited data (76 patients), restricting the depth of model training. The release of the RSNA Intracranial Hemorrhage dataset~\cite{RSNA2020} marked a significant milestone, offering over 25,000 studies; however, it only provides slice-level binary labels, lacking the bounding box annotations necessary for precise localization tasks. Similarly, while the CQ500 dataset~\cite{CQ5002018} is widely used for validation, it lacks the pixel-level or bounding-box annotations required for supervising localization models. Recent efforts such as the PHE–SICH-CT-IDS dataset~\cite{ma2024phe} have introduced high-quality segmentation benchmarks, yet they remain limited in scale (120 patients). In this study, the Hemorica dataset~\cite{Hemorica2025} is utilized. With 372 patients and precise segmentations that can be converted to bounding boxes, it offers a significantly larger and more suitable benchmark for training robust object-level tracking systems.

\subsection{Detection Architectures}
Historically, ICH detection relied on 2D Convolutional Neural Networks (CNNs) that treated each slice as an independent image. While computationally efficient, these models inherently lack volumetric context. To address this amnesia, researchers introduced sequence modeling. Burduja \textit{et al.} \cite{Burduja2020ICH} combined a 2D CNN with an LSTM to aggregate features across slices, achieving top-tier performance in the RSNA challenge. Similarly, Ngo \textit{et al.} \cite{Ngo2022ICH} utilized deep descriptors of adjacent slices to stabilize classification.

The field subsequently pivoted toward fully volumetric approaches. Ye \textit{et al.} \cite{ref5} proposed a 3D Joint CNN-RNN framework to capture spatial continuity, while recent work by Subramanian \textit{et al.} \cite{ref6} utilized U-shaped 3D processing models for precise subtype segmentation. However, these volumetric methods demand massive GPU memory, creating a barrier to deployment in resource-constrained clinical settings. The proposed framework targets the specific operational gap left by these methods: achieving volumetric consistency without the prohibitive hardware costs of 3D CNNs. By replacing heavy feature-aggregation modules with lightweight motion estimation, 3D-aware inference is enabled on edge devices where standard volumetric models cannot deploy.

Furthermore, the architectural choice between classification, segmentation, and detection is often dictated by the trade-off between supervision cost and interpretability. Volumetric segmentation models require dense voxel-level annotations, which are scarce and costly. On the other hand, classification models suffer from opacity, necessitating secondary tools to locate the pathology. Rafati \textit{et al.}~\cite{rafati2025benchmarking} addressed this by benchmarking different CAM methods on the Hemorica dataset to assess how classification models can be interpreted. However, object detection bridges this gap by employing bounding-box supervision. This method is significantly faster to annotate than segmentation masks while still providing explicit spatial supervision, ensuring the model learns to identify the specific region of the hemorrhage rather than relying on global image statistics.

\subsection{Object Tracking in Medical Imaging}
Tracking in medical imaging has traditionally referred to two distinct tasks: longitudinal monitoring (tracking a lesion's growth over months) and dynamic organ tracking (tracking a beating heart). Cai \textit{et al.} \cite{Cai2021DeepLesion} introduced the \textit{Deep Lesion Tracker} to match lesions across 4D longitudinal studies, while Yan \textit{et al.} \cite{Yan2018DeepLesion} established the DeepLesion benchmark to facilitate large-scale lesion mining. In dynamic imaging, Yu \textit{et al.} \cite{Yu2022Polyp} integrated an instance tracking head into a polyp detector for colonoscopy videos, and Lei \textit{et al.} \cite{Lei2024Cardiac} applied tracking for real-time cardiac ultrasound guidance.

This work introduces a third category: Slice-to-Slice Lesion Tracking. Video object trackers are adapted to static 3D volumes. Specifically, ByteTrack~\cite{zhang2022bytetrack}, a Multi-Object Tracker (MOT), is leveraged. Unlike prior algorithms such as SORT~\cite{bewley2016simple} or DeepSORT~\cite{wojke2017simple} that discard low-confidence detections, ByteTrack utilizes a two-stage matching process to recover weak detections. The proposed approach diverges from standard medical tracking by treating the static $z$-axis as a temporal stream. The framework specifically leverages ByteTrack's ability to associate low-confidence detections a critical feature for identifying the faint, fuzzy boundaries of hemorrhage that are typically discarded by strict thresholding in standard tracking algorithms.

\section{Methodology}
The proposed approach fundamentally reinterprets the problem of Intracranial Hemorrhage (ICH) detection. Instead of treating a CT scan as a stack of unrelated images, the $z$-axis is treated as a temporal dimension, effectively converting the 3D volume into a video sequence. This enables the application of Multi-Object Tracking (MOT) techniques to recover lesions that might be missed by a standalone detector.

\subsection{Data Description and Preprocessing}
The Hemorica dataset, a multi-institutional collection of non-contrast head CT examinations, was utilized. The dataset comprises 327 patients, totaling 12,067 axial slices. Of these, 2,679 slices are labeled as hemorrhage-positive, while the remaining 9,388 are negative controls. The positive cases encompass five distinct subtypes: Intracerebral (ICH), Intraventricular (IVH), Epidural (EPH), Subdural (SDH), and Subarachnoid (SAH) hemorrhages.

A significant challenge in this domain is class imbalance; negative slices account for over 75\% of the dataset, and certain subtypes like Epidural Hemorrhage are rare (approx. 1.3\% of total slices). To address this and improve model robustness, a binary classification scheme was adopted, aggregating all subtypes into a single Hemorrhage class. 

For model development, a patient-level split was strictly enforced to prevent data leakage, ensuring that no slices from a training patient appear in the test set. A stratified 80/20 split was utilized: 80\% of studies (261 patients) were reserved for training, and 20\% (66 patients) were set aside for independent testing. 
To prepare the data for the network, a standard brain window (Level: 40, Width: 80) was applied to all DICOM series. This windowing technique highlights coagulated blood while suppressing bone artifacts and soft tissue noise.

\subsection{2D Baseline Detection}
The selection of the primary detection architecture was governed by the strict latency requirements 
of medical triage systems deployed at the edge. While larger model variants offer higher parametric 
capacity, we focused exclusively on the Nano configurations of the YOLO family: 
YOLOv8 \cite{yolov8_ultralytics}, YOLOv10 \cite{Wang_2024_YOLOv10}, 
YOLOv11 \cite{yolo11_ultralytics}, and the recently released YOLOv12 \cite{Tian_2025_YOLOv12}. 
The model was trained on individual 2D slices to learn the visual features of hemorrhage. The model was explicitly trained without any data augmentation (no rotation, scaling, or mosaic). This design choice prioritizes the preservation of anatomical integrity; unlike natural images, medical scans possess strict structural consistency, and heavy geometric distortions risk introducing synthetic artifacts that could compromise feature learning. By training on clean data, a pure baseline was established, allowing for the attribution of any subsequent performance gains strictly to the tracking logic.

\subsection{Deep Multi-Object Tracking (ByteTrack)}
While YOLO provides candidate detections, slice-independent detectors inherently lack temporal consistency, often leading to intermittent false negatives (flickering) across the $z$-axis. To enforce consistency, ByteTrack was integrated. Unlike traditional trackers that discard weak detections, ByteTrack utilizes a two-stage matching process identified as critical for recovering faint hemorrhages:
\begin{enumerate}
    \item \textit{High-Confidence Matching:} First, boxes with high detection scores are associated with existing tracks using the Kalman Filter to predict the lesion's next position.
    \item \textit{Low-Confidence Recovery:} ByteTrack keeps weak detections (which are often ignored) and attempts to match them to existing tracks using Intersection over Union (IoU). This step facilitates the recovery of hemorrhages that are partially obscured or visually subtle in a specific slice.
\end{enumerate}

\subsection{Bi-directional Tracking Strategy}
Standard online trackers utilizing Kalman filters inherently require a strictly causal sequence to initialize state covariance, often resulting in a warm-up lag. In the context of CT volumes, where a hemorrhage may present immediately in the initial slices, this latency creates a risk of missed detections. 
To mitigate this limitation, a \textit{bi-directional tracking} module was implemented. Every CT volume is processed twice:
\begin{itemize}
    \item \textit{Forward Pass ($1 \rightarrow N$):} Tracks lesions from the skull base to the vertex.
    \item \textit{Backward Pass ($N \rightarrow 1$):} Tracks lesions in reverse order.
\end{itemize}
The final set of tracked lesions is the union of these two passes. This ensures that a lesion missed during the initialization phase of the forward pass is successfully captured as a stable track during the backward pass.

\subsection{Hybrid Inference and Refinement}
Relying solely on the tracker can sometimes suppress isolated but obvious findings. To prevent this, a \textit{hybrid inference} strategy was employed. All High Confidence YOLO detections ($Confidence > 0.2$) are retained regardless of whether the tracker linked them. This acts as a safety net, ensuring that distinct, high-probability lesions are never discarded.

\subsection{Spatiotemporal Consistency Filtering}
Finally, to differentiate transient noise from true pathology without the complexity of state estimation, a simplified, rule-based filter was formulated. Recognizing that the standard Kalman filter used in ByteTrack introduces computational overhead and initialization latency, a direct spatial association method was chosen. It was hypothesized that for stationary anatomical structures, complex motion prediction is unnecessary; mere spatial overlap between adjacent slices is a sufficient proxy for volumetric continuity.

Therefore, a spatiotemporal consistency filter was implemented that operates solely on Intersection over Union (IoU). For every candidate bounding box in slice $z$, its spatial alignment is verified with detections in the preceding slice ($z-1$) and the succeeding slice ($z+1$). The logic dictates that a true volumetric lesion must exhibit physical continuity. Consequently, if a detection fails to overlap (IoU $> 0$) with any region in \textit{either} of its neighboring slices, it is classified as isolated noise and eliminated. This approach reduces the tracking mechanism to its most essential component geometric overlap ensuring high precision without the warm-up lag or computational cost of predictive filters.

\subsection{Evaluation Metrics}
To assess the performance of the detection pipeline, standard object detection metrics are utilized: Precision, Recall, and the F1-score. A detection is considered a True Positive (TP) if the Intersection over Union (IoU) between the predicted bounding box and the ground truth mask exceeds a threshold of 0.5. 
\begin{itemize}
    \item \textit{Precision} measures the reliability of positive predictions ($TP / (TP + FP)$). In a clinical setting, high precision reduces false alarms, which prevents radiologist fatigue.
    \item \textit{Recall (Sensitivity)} measures the proportion of actual hemorrhages correctly identified ($TP / (TP + FN)$). This is the most critical metric for triage, as missing a hemorrhage can have fatal consequences.
    \item \textit{F1-Score} is the harmonic mean of Precision and Recall, providing a single metric to evaluate the balance between false alarms and missed cases.
\end{itemize}

\section{Experiments and Results}

To assess the proposed framework, slice-level performance was evaluated using Precision, Recall, and the F1-score. Given the critical nature of Intracranial Hemorrhage detection, the primary objective was to maximize Recall to minimize the risk of missed diagnoses, while simultaneously maintaining high Precision to prevent alert fatigue. The F1-score served as the global metric for balancing these competing goals.

\begin{table}[htbp]
\caption{Benchmark of YOLO Nano Architectures for Backbone Selection}
\label{tab:backbone}
\centering
\begin{tabular}{l S[table-format=1.2] S[table-format=1.1] S[table-format=1.3] S[table-format=1.3]}
\toprule
\textbf{Model} & \textbf{Params(M)} & \textbf{FLOPs(G)} & \textbf{Recall} & \textbf{mAP@50} \\
\midrule
YOLOv8n  & 3.2 & 8.7 & 0.537 & 0.595 \\
YOLOv10n & 2.3 & 6.7 & 0.509 & 0.594 \\
\textbf{YOLOv11n} & \textbf{2.6} & \textbf{6.5} & \textbf{0.542} & \textbf{0.631} \\
YOLOv12n & 2.6 & 6.5& 0.529 & 0.597 \\
\bottomrule
\end{tabular}
\end{table}

\begin{table*}[!t]
\caption{Ablation Study of Methods on the \textbf{Training Set}}
\label{tab:train_results}
\centering
\begin{tabular}{l c c c S[table-format=1.3] S[table-format=1.3] S[table-format=1.3]}
\toprule
\textbf{Method} & \textbf{Track Act.} & \textbf{Min Match} & \textbf{Lost Buff.} & {\textbf{Precision}} & {\textbf{Recall}} & {\textbf{F1-score}} \\
\midrule
Baseline YOLOv11n       & n/a & n/a & n/a & 0.970 & 0.979 & 0.974 \\
ByteTrack           & 0.35 & 0.95 & 5   & 0.999 & 0.541 & 0.702 \\
BiDirectional       & 0.35 & 0.95 & 5   & 0.998 & 0.713 & 0.832 \\
\textbf{Hybrid ByteTrack}    & 0.35 & 0.95 & 5   & \textbf{0.987} & \textbf{0.979} & \textbf{0.974} \\
Spatiotemporal Filter & n/a & n/a & n/a & 0.970 & 0.979 & 0.974 \\
\bottomrule
\end{tabular}
\end{table*}

\begin{table*}[!t]
\caption{Performance on the \textbf{Test Set} (Unseen Patients)}
\label{tab:test_results}
\centering
\begin{tabular}{l c c c S[table-format=1.3] S[table-format=1.3] S[table-format=1.3]}
\toprule
\textbf{Method} & \textbf{Track Act.} & \textbf{Min Match} & \textbf{Lost Buff.} & {\textbf{Precision}} & {\textbf{Recall}} & {\textbf{F1-score}} \\
\midrule
Baseline YOLOv11n       & n/a & n/a & n/a & 0.703 & 0.643 & 0.674 \\
ByteTrack           & 0.35 & 0.95 & 5   & 0.969 & 0.376 & 0.542 \\
BiDirectional       & 0.35 & 0.95 & 5   & 0.965 & 0.482 & 0.643 \\
\textbf{Hybrid ByteTrack}    & 0.35 & 0.95 & 5   & \textbf{0.779} & \textbf{0.647} & \textbf{0.707} \\
Spatiotemporal Filter & n/a & n/a & n/a & 0.722 & 0.640 & 0.679 \\
\bottomrule
\end{tabular}
\end{table*}

\subsection{Backbone Architecture Comparison}
To identify the most efficient 2D backbone, we benchmarked four generations of the 
YOLO Nano family. As summarized in Table~\ref{tab:backbone}, the choice was driven 
by the trade-off between localization accuracy ($mAP_{50}$) and computational 
efficiency (GFLOPs). We compared the established YOLOv8n \cite{yolov8_ultralytics}, 
the NMS-free YOLOv10n \cite{Wang_2024_YOLOv10}, the optimized YOLOv11n \cite{yolo11_ultralytics}, 
and the attention-centric YOLOv12n \cite{Tian_2025_YOLOv12}. \textbf{YOLOv11n} was 
selected as the optimal primary detector as it achieved the highest $mAP_{50}$.

\subsection{Experimental Setup}
All models were implemented in PyTorch and trained on an NVIDIA Tesla P100 GPU. The YOLOv11n backbone was trained for 50 epochs with a batch size of 16 using the AdamW optimizer. To ensure reproducibility and isolate the impact of the tracking logic, all data augmentation (mosaic, scaling, rotation) was disabled.

For the tracking modules, a comprehensive grid search was conducted to optimize key hyperparameters. The following parameters were evaluated:
\begin{itemize}
    \item Track Activation Thresholds: 0.20 to 1.0 (step 0.05)
    \item Minimum Matching Thresholds: 0.50 to 1.0 (step 0.05)
    \item Lost Track Buffer sizes: \{3, 5, 7, 9\}
\end{itemize}
Based on this sweep, the optimal configuration for the reported results was determined to be: Track Activation = 0.35, Minimum Matching = 0.95, and Buffer = 5.

\subsection{Training Dynamics}
\begin{figure}[!t]
\centering
\includegraphics[width=\columnwidth]{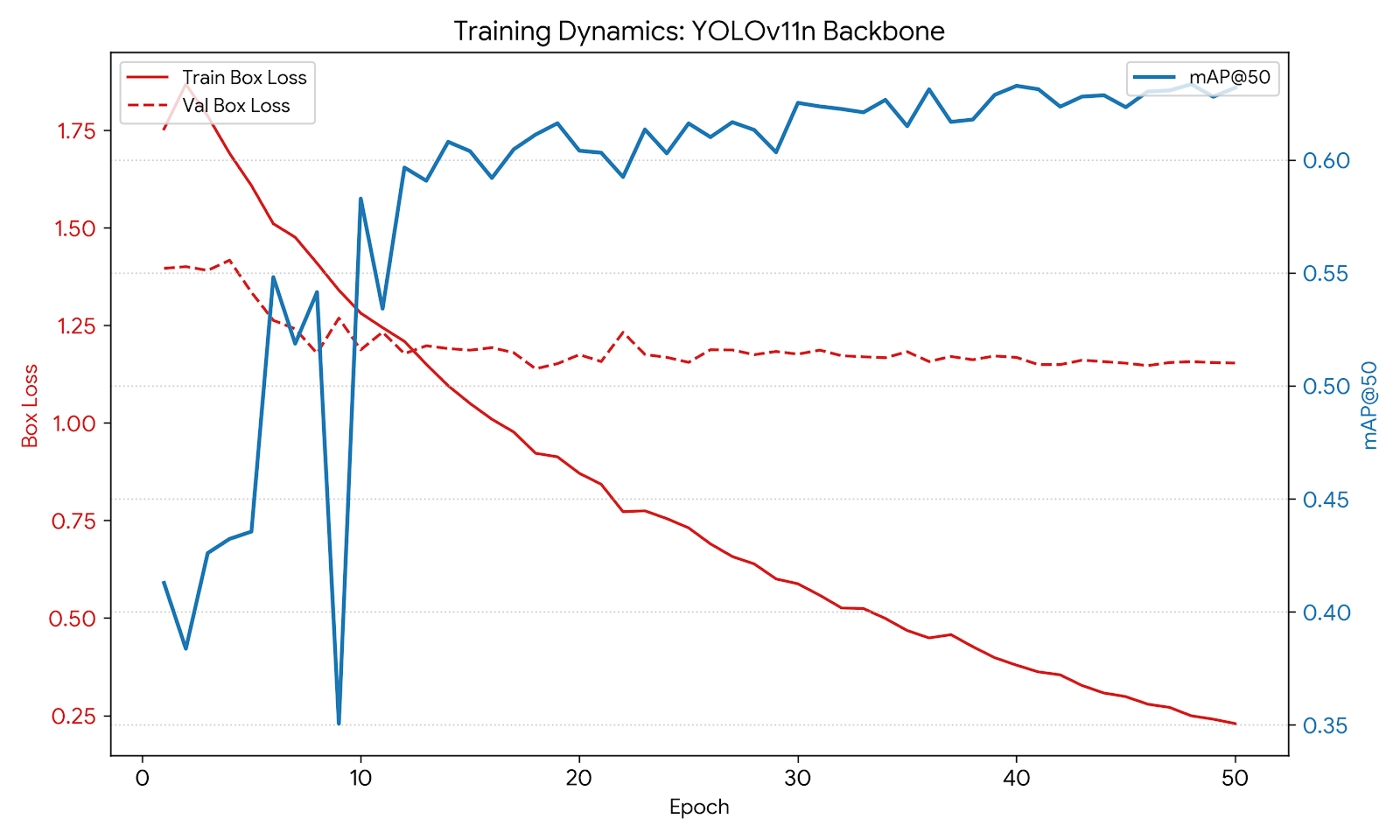}
\caption{\textbf{Training Dynamics.} Evolution of Box Loss (Red) and mAP@50 (Blue) over 50 epochs. The validation loss closely tracks the training loss, confirming stable convergence without overfitting.}
\label{fig:training_curves}
\end{figure}

Before integrating the temporal tracking module, the stability of the baseline 2D detector was verified. As illustrated in Fig.~\ref{fig:training_curves}, the YOLOv11n backbone demonstrates consistent convergence. The validation box loss (dashed red line) tracks the training loss (solid red line) closely, indicating that the model successfully learned feature representations without overfitting. Concurrently, the mAP@50 rises steadily, plateauing around epoch 45.

\subsection{Hyperparameter Optimization}
To ensure the detector operated at its optimal point prior to tracking, a hyperparameter sweep was performed on the training set. Confidence thresholds ranging from 0.05 to 0.80 were evaluated. As detailed in Table~\ref{tab:threshold_tuning}, performance peaks at a threshold of \textbf{0.20} (F1 = 0.946). Thresholds below 0.10 yielded marginally higher recall but introduced excessive noise, while values above 0.30 aggressively suppressed true positive findings. Consequently, a confidence threshold of 0.20 was fixed for all subsequent experiments.

\begin{table}[!t]
\caption{Threshold Optimization (Baseline YOLO)}
\label{tab:threshold_tuning}
\centering
\begin{tabular}{S[table-format=1.2] S[table-format=1.3] S[table-format=1.3] S[table-format=1.3]}
\toprule
{\textbf{Threshold}} & {\textbf{Precision}} & {\textbf{Recall}} & {\textbf{F1-Score}} \\
\midrule
0.05 & 0.958 & 0.908 & 0.932 \\
0.10 & 0.978 & 0.907 & 0.941 \\
\textbf{0.20} & \textbf{0.994} & \textbf{0.902} & \textbf{0.946} \\
0.30 & 0.997 & 0.898 & 0.945 \\
0.40 & 0.998 & 0.893 & 0.943 \\
0.50 & 0.999 & 0.881 & 0.937 \\
0.60 & 0.999 & 0.866 & 0.928 \\
0.70 & 0.999 & 0.845 & 0.916 \\
0.80 & 0.999 & 0.768 & 0.869 \\
\bottomrule
\end{tabular}
\end{table}

\subsection{Quantitative Results}
The impact of the temporal post-processing modules was analyzed on both the Training Set (Table~\ref{tab:train_results}) and the independent Test Set (Table~\ref{tab:test_results}).

The Baseline 2D model achieved a strong Recall of 0.643 on the test set but suffered from low Precision (0.703), indicating frequent false positives. Applying standard ByteTrack boosted Precision to 0.969 but caused a drop in Recall (to 0.376) due to the warm-up lag.

\begin{figure*}[!t]
\centering
\includegraphics[width=\textwidth]{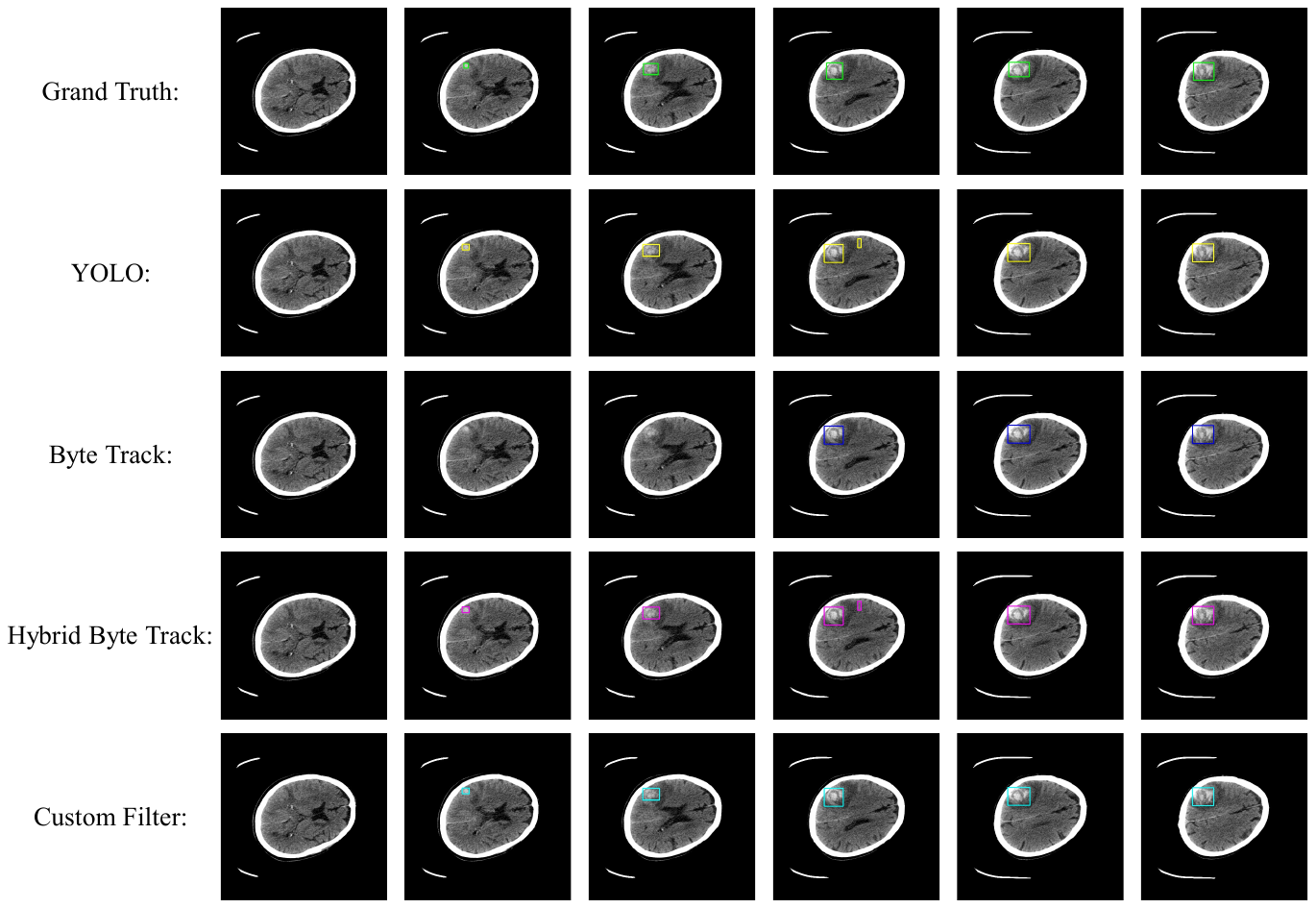}
\caption{\textbf{Qualitative comparison across methods.} 
We visualize consecutive slices (Columns) to demonstrate temporal consistency.
\textbf{Row 1 (Green):} Ground Truth annotations.
\textbf{Row 2 (Yellow):} Baseline YOLOv11n detections showing slice-level inconsistencies.
\textbf{Row 3 (Blue):} ByteTrack results, showing track initialization lag.
\textbf{Row 4 (Purple):} Proposed Hybrid method.
\textbf{Row 5 (Teal):} Spatiotemporal Filter results.
Note how the proposed methods (Rows 4-5) recover the missed detections in the middle columns compared to the baseline.}
\label{fig:qualitative}
\end{figure*}

The \textbf{Hybrid ByteTrack} strategy successfully resolved this trade-off. By fusing high-confidence YOLO detections with the tracker's associations, the system maintained the high Recall of the baseline (0.643 vs 0.647) while significantly improving Precision (0.703 vs 0.779). This resulted in the highest overall F1-score of 0.707. These quantitative results demonstrate that the primary value of tracking is serving as a temporal validator filtering out inconsistent 2D noise while preserving the detector's native sensitivity.

\subsection{Qualitative Assessment}
Visual analysis further confirms these findings. As shown in Fig.~\ref{fig:qualitative}, the standard tracking approach (Blue) fails to generate bounding boxes for the initial slices of the lesion due to the initialization lag. In contrast, the Hybrid output (Purple) successfully retains these early detections. Notably, it was observed that the baseline detector was successful in detection of the central slices which were large hemorrhages; the primary contribution of the pipeline was resolving random noises that were mistaken with the lesion and rejecting those false positives in healthy tissue.

\section{Discussion}
The central hypothesis of this study was that treating CT scans as video sequences would recover missed hemorrhages. However, the quantitative results reveal a more nuanced reality. The pure ByteTrack experiment demonstrated that while tracking introduces temporal consistency, it initially harms sensitivity due to initialization lag. The success of the Hybrid method suggests that the primary value of tracking in this domain is not necessarily discovering new lesions that the detector missed, but rather acting as a rigorous temporal validator. By suppressing isolated false positives (boosting Precision from 0.703 to 0.779) while retaining the detector's high-confidence findings, the system effectively mimics the cognitive process of a radiologist: trusting a strong visual signal immediately, but requiring contextual validation for ambiguous ones.

This finding has significant implications for deployment in resource-constrained environments. A key motivation for this work was the computational bottleneck of 3D processing models. The results demonstrate that 3D contextual reasoning can be approximated using purely 2D tools. By chaining a lightweight YOLO detector with a Kalman Filter, volumetric consistency is achieved without the massive VRAM overhead of 3D convolutions. This confirms that the $z$-axis of a CT scan contains predictable motion dynamics that can be exploited by standard video algorithms, provided the warm-up and boundary issues are addressed via the Bi-directional and Hybrid logic.

Furthermore, a comparison between the Training Set (Table~\ref{tab:train_results}) and Test Set (Table~\ref{tab:test_results}) highlights the inherent challenge of medical generalization. On the training data, where the detector has learned the specific texture of the hemorrhages, the Hybrid Tracker achieves nearly perfect performance (F1=0.974). The drop in performance on the unseen Test Set (F1=0.707) indicates that inter-patient variability remains a dominant hurdle. However, crucially, the relative improvement provided by the tracking module remains consistent across both sets. This suggests that while the underlying detector's feature extraction may degrade on unseen patients, the \textit{logic} of the video-viewpoint framework is robust and transferable.

A limitation of the current approach is its reliance on spatial overlap (IoU) for association. If a patient moves significantly between slices or if the lesion shifts rapidly, the Kalman Filter may lose the track. Future integration of appearance-based Re-Identification (ReID) features could resolve this by allowing the system to visually match a lesion across a gap, rather than relying solely on spatial coordinates.

\section{Conclusion}
In this work, a video-viewpoint framework for Intracranial Hemorrhage detection was introduced, shifting the paradigm from static slice analysis to dynamic lesion tracking. By adapting the ByteTrack algorithm with a Hybrid inference strategy, the initialization lag inherent in video trackers was successfully overcome. The results demonstrate that this approach enhances diagnostic precision (from 0.703 to 0.779) by eliminating non-volumetric noise, offering a computationally efficient alternative to heavy 3D architectures.

The proposed  system addresses the critical diagnostic bottleneck in remote and after-hours clinics, providing a lightweight, high-precision triage tool that runs on standard hardware. Future work will focus on closing the generalization gap through domain-adaptive training and exploring the BoT-SORT framework to leverage visual ReID features for more robust occlusion handling.

\bibliographystyle{ieeetr}
\bibliography{references}

@article{Burduja2020ICH,
  author       = {Burduja, Mihai and Ionescu, Radu Tudor and Verga, Nicolae},
  title        = {Accurate and Efficient Intracranial Hemorrhage Detection and Subtype Classification in 3D CT Scans with Convolutional and Long Short‐Term Memory Neural Networks},
  journal      = {Sensors},
  year         = {2020},
  volume       = {20},
  number       = {19},
  pages        = {5611},
  doi          = {10.3390/s20195611}
}

@article{Ngo2022ICH,
  author    = {Ngo, Dat T. and Nguyen, Thao T. B. and Nguyen, Hieu T. and others},
  title     = {Slice‐level Detection of Intracranial Hemorrhage on CT Using Deep Descriptors of Adjacent Slices},
  journal   = {arXiv preprint arXiv:2208.03403},
  year      = {2022}
}

@article{ref5,
  author       = {Ye, Haipeng and Gao, Fei and Yin, Yilong and others},
  title        = {Precise diagnosis of intracranial hemorrhage and subtypes using a three-dimensional joint convolutional and recurrent neural network},
  journal      = {European Radiology},
  year         = {2019},
  volume       = {29},
  number       = {11},
  pages        = {6191--6201},
  doi          = {10.1007/s00330-019-06163-2}
}

@article{ref6,
  author       = {Subramanian, Bargava and Kumarasami, Naveen and Shastry, Praveen and others},
  title        = {3D Convolutional Neural Networks for Improved Detection of Intracranial bleeding in CT Imaging},
  journal      = {arXiv preprint arXiv:2503.20306},
  year         = {2025}
}

@article{RSNA2020,
  author = {Flanders, Adam E. and Prevedello, Luciano M. and Shih, George and Halabi, Safwan S. and Kalpathy-Cramer, Jayashree and Ball, Robyn and Mongan, John T. and Stein, Anouk and Kitamura, Felipe C. and Lungren, Matthew P. and Cossa, Chris and Colakoğlu, Errol}, 
  title = {Construction of a Machine Learning Dataset through Collaboration: The RSNA 2019 Brain CT Hemorrhage Challenge},
  journal = {Radiology: Artificial Intelligence},
  year = {2020},
  volume = {2},
  number = {3},
  pages = {e190217},
  doi = {10.1148/ryai.2020190217}
}

@article{CQ5002018,
  title        = {Development and Validation of Deep Learning Algorithms for Detection of Critical Findings in Head CT Scans},
  author       = {Chilamkurthy, Sasank and Ghosh, Ranjan and Tanamala, Harsha and others},
  journal      = {arXiv preprint arXiv:1803.05854},
  year         = {2018}
}

@inproceedings{zhang2022bytetrack,
  title        = {ByteTrack: Multi-object Tracking by Associating Every Detection Box},
  author       = {Zhang, Yifu and Sun, Peize and Jiang, Yi and others},
  booktitle    = {Proceedings of the European Conference on Computer Vision (ECCV)},
  pages        = {1--21},
  year         = {2022}
}

@article{Hemorica2025,
  title        = {Hemorica: A Comprehensive CT Scan Dataset for Automated Brain Hemorrhage Classification, Segmentation, and Detection},
  author       = {Davoodi, Kasra and Hoseyni, Mohammad and Khoramdel, Javad and Barati, Reza and Mortazavi, Reihaneh and Nikoofard, Amirhossein and Aliyari-Shoorehdeli, Mahdi and Parikhan, Jaber Hatam},
  journal      = {arXiv preprint arXiv:2509.22993},
  year         = {2025}
}

@inproceedings{bewley2016simple,
  title        = {Simple online and realtime tracking},
  author       = {Bewley, Alex and Ge, Zongyuan and Ott, Lionel and Ramos, Fabio and Upcroft, Ben},
  booktitle    = {2016 IEEE International Conference on Image Processing (ICIP)},
  pages        = {3464--3468},
  year         = {2016}
}

@inproceedings{wojke2017simple,
  title        = {Simple online and realtime tracking with a deep association metric},
  author       = {Wojke, Nicolai and Bewley, Alex and Paulus, Dietrich},
  booktitle    = {2017 IEEE International Conference on Image Processing (ICIP)},
  pages        = {3645--3649},
  year         = {2017}
}

@inproceedings{Cai2021DeepLesion,
  title        = {Deep lesion tracker: Monitoring lesions in 4d longitudinal imaging studies},
  author       = {Cai, Jinzheng and Yan, Ke and Lu, Le and others},
  booktitle    = {Proceedings of the IEEE/CVF Conference on Computer Vision and Pattern Recognition (CVPR)},
  pages        = {15159--15169},
  year         = {2021}
}

@article{Yan2018DeepLesion,
  title        = {DeepLesion: automated mining of large-scale lesion annotations and universal lesion detection with deep learning},
  author       = {Yan, Ke and Wang, Xiaosong and Lu, Le and Summers, Ronald M},
  journal      = {Journal of Medical Imaging},
  volume       = {5},
  number       = {3},
  pages={036501},
  year={2018}
}

@article{Yu2022Polyp,
  title        = {An end-to-end tracking method for polyp detectors in colonoscopy},
  author       = {Yu, T and others},
  journal      = {Artificial Intelligence in Medicine},
  volume       = {131},
  pages        = {102363},
  year         = {2022}
}

@inproceedings{Lei2024Cardiac,
  title        = {Epicardium prompt-guided real-time cardiac ultrasound},
  author       = {Lei, Y and others},
  booktitle    = {Medical Image Computing and Computer Assisted Intervention – MICCAI 2024},
  year         = {2024}
}

@article{greenberg2022guideline,
  title={2022 Guideline for the Management of Patients With Spontaneous Intracerebral Hemorrhage: A Guideline From the American Heart Association/American Stroke Association},
  author={Greenberg, Steven M and Ziai, Wendy C and Cordonnier, Charlotte and Dowlatshahi, Dar and Francis, Brian and Goldstein, Larry N and Hemphill, J Claude and Johnson, Reed and Keigher, Kiffon M and Mack, William J and others},
  journal={Stroke},
  volume={53},
  number={7},
  pages={e282--e361},
  year={2022},
  publisher={American Heart Association}
}

@article{hssayeni2020intracranial,
  title={Intracranial hemorrhage segmentation using a deep convolutional model},
  author={Hssayeni, Murtadha D and Croock, Muayad S and Salman, Aymen D and Al-Khafaji, Hassan Falah and Yahya, Zakaria A and Ghoraani, Behnaz},
  journal={Data},
  volume={5},
  number={1},
  pages={14},
  year={2020},
  publisher={MDPI}
}

@article{ma2024phe,
  title={Phe-sich-ct-ids: A benchmark ct image dataset for evaluation semantic segmentation, object detection and radiomic feature extraction of perihematomal edema in spontaneous intracerebral hemorrhage},
  author={Ma, Deguo and Li, Chen and Du, Tianming and Qiao, Lin and Tang, Dechao and Ma, Zhiyu and Shi, Liyu and Lu, Guotao and Meng, Qingtao and Chen, Zhihao and others},
  journal={Computers in Biology and Medicine},
  volume={173},
  pages={108342},
  year={2024},
  publisher={Elsevier}
}

@article{rafati2025benchmarking,
  title={Benchmarking Class Activation Map Methods for Explainable Brain Hemorrhage Classification on Hemorica Dataset},
  author={Rafati, Zahra and Hoseyni, Mohammad and Khoramdel, Javad and Nikoofard, Amirhossein},
  journal={arXiv preprint arXiv:2508.17699},
  year={2025}
}

@article{Larson2011,
  title={National trends in CT use in the emergency department: 1995--2007},
  author={Larson, David B and Johnson, Lara W and Schnell, Beverly M and Salisbury, Shelia R and Forman, Howard P},
  journal={Radiology},
  volume={258},
  number={1},
  pages={164--173},
  year={2011},
  publisher={RSNA}
}

@article{Coles2007,
  title={Imaging after brain injury},
  author={Coles, Jonathan P},
  journal={British Journal of Anaesthesia},
  volume={99},
  number={1},
  pages={49--60},
  year={2007},
  publisher={Oxford University Press}
}

@article{Papa2012,
  title={Performance of the Canadian CT head rule and the New Orleans criteria for predicting any traumatic intracranial injury on computed tomography in a United States level I trauma center},
  author={Papa, Linda and Stiell, Ian G and Clement, Catherine M and Pawlowicz, Artur and Wolfram, Andrew and Braga, Carolina and Draviam, Sameer and Wells, George A},
  journal={Academic Emergency Medicine},
  volume={19},
  number={1},
  pages={2--10},
  year={2012},
  publisher={Wiley Online Library}
}

@article{Wysoki1998,
  title={Head trauma: CT scan interpretation by radiology residents versus staff radiologists},
  author={Wysoki, Michael G and Nassar, Carlos J and Koenigsberg, Robert A and Novelline, Robert A and Faro, Scott H and Faerber, Eric N},
  journal={Radiology},
  volume={208},
  number={1},
  pages={125--128},
  year={1998},
  publisher={RSNA}
}

@article{Erly2002,
  title={Radiology resident evaluation of head CT scan orders in the emergency department},
  author={Erly, William K and Berger, William G and Krupinski, Elizabeth and Seeger, Joachim F and Guisto, John A},
  journal={American Journal of Neuroradiology},
  volume={23},
  number={1},
  pages={103--107},
  year={2002},
  publisher={Am Soc Neuroradiology}
}

@article{Gao2017,
  title={Classification of CT brain images based on deep learning networks},
  author={Gao, Xiaohong W and Hui, Rui and Tian, Zengmin},
  journal={Computer methods and programs in biomedicine},
  volume={138},
  pages={49--56},
  year={2017},
  publisher={Elsevier}
}

@article{Grewal2017,
  title={RADNET: Radiologist level accuracy using deep learning for hemorrhage detection in CT scans},
  author={Grewal, Monika and Srivastava, Muktabh Mayank and Kumar, Pulkit and Varadarajan, Srikrishna},
  journal={arXiv preprint arXiv:1710.04934},
  year={2017}
}

@article{Litjens2017,
  title={A survey on deep learning in medical image analysis},
  author={Litjens, Geert and Kooi, Thijs and Bejnordi, Babak Ehteshami and Setio, Arnaud Arindra Adiyoso and Ciompi, Francesco and Ghafoorian, Mohsen and van der Laak, Jeroen AWM and van Ginneken, Bram and S{\'a}nchez, Clara I},
  journal={Medical image analysis},
  volume={42},
  pages={60--88},
  year={2017},
  publisher={Elsevier}
}

@inproceedings{Redmon_2016_CVPR,
  author    = {Redmon, Joseph and Divvala, Santosh and Girshick, Ross and Farhadi, Ali},
  title     = {You Only Look Once: Unified, Real-Time Object Detection},
  booktitle = {Proceedings of the IEEE Conference on Computer Vision and Pattern Recognition (CVPR)},
  year      = {2016},
  pages     = {779-788},
  doi       = {10.1109/CVPR.2016.91}
}

@misc{yolo11_ultralytics,
  author  = {Glenn Jocher and Jing Qiu},
  title   = {Ultralytics YOLO11},
  version = {11.0.0},
  year    = {2024},
  url     = {https://github.com/ultralytics/ultralytics},
  note    = {Available at https://github.com/ultralytics/ultralytics}
}

@article{Wang_2024_YOLOv10,
  title   = {YOLOv10: Real-Time End-to-End Object Detection},
  author  = {Wang, Ao and Chen, Hui and Liu, Lihao and Chen, Kai and Lin, Zijia and Han, Jungong and Ding, Guiguang},
  journal = {arXiv preprint arXiv:2405.14458},
  year    = {2024}
}

@misc{yolov8_ultralytics,
  author  = {Glenn Jocher and Ayush Chaurasia and Jing Qiu},
  title   = {Ultralytics YOLOv8},
  version = {8.0.0},
  year    = {2023},
  url     = {https://github.com/ultralytics/ultralytics},
  note    = {Available at https://github.com/ultralytics/ultralytics}
}

@article{Tian_2025_YOLOv12,
  title   = {YOLO12: Attention-Centric Real-Time Object Detectors},
  author  = {Tian, Yunjie and Ye, Qixiang and Doermann, David},
  journal = {arXiv preprint arXiv:2502.12524},
  year    = {2025}
}

\end{document}